\pdfoutput=1

\documentclass[11pt]{article}

\usepackage[final]{acl}

\usepackage{times}
\usepackage{latexsym}

\usepackage[T1]{fontenc}

\usepackage[utf8]{inputenc}

\usepackage{microtype}

\usepackage{inconsolata}

\usepackage{graphicx}

\usepackage{amsmath}
\usepackage{amsfonts}
\usepackage{subcaption}
\usepackage{mwe}
\usepackage{multirow}
\usepackage{booktabs}
\usepackage{tabularx}

\title{Position Encoding with Random Float Sampling Enhances\\Length Generalization of Transformers}

\author{Atsushi Shimizu\thanks{Work done as a research intern at the University of Tokyo.} \\
  Daiwa Securities \\
  \texttt{atsushi.shimizu@daiwa.co.jp} \\\And
  Shohei Taniguchi \\
  The University of Tokyo \\
  \texttt{taniguchi@weblab.t.u-tokyo.ac.jp} \\\AND
  Yutaka Matsuo \\
  The University of Tokyo \\
  \texttt{matsuo@weblab.t.u-tokyo.ac.jp} \\}

\begin{document}
\maketitle
\begin{abstract}
Length generalization is the ability of language models to maintain performance on inputs longer than those seen during pretraining. In this work, we introduce a simple yet powerful position encoding (PE) strategy, Random Float Sampling (RFS), that generalizes well to lengths unseen during pretraining or fine-tuning. In particular, instead of selecting position indices from a predefined discrete set, RFS uses randomly sampled continuous values, thereby avoiding out-of-distribution (OOD) issues on unseen lengths by exposing the model to diverse indices during training. Since assigning indices to tokens is a common and fundamental procedure in widely used PEs, the advantage of RFS can easily be incorporated into, for instance, the absolute sinusoidal encoding, RoPE, and ALiBi. Experiments corroborate its effectiveness by showing that RFS results in superior performance in length generalization tasks as well as zero-shot commonsense reasoning benchmarks.
\end{abstract}

\section{Introduction}

\begin{figure}[t]
    \centering
    \includegraphics[width=\columnwidth]{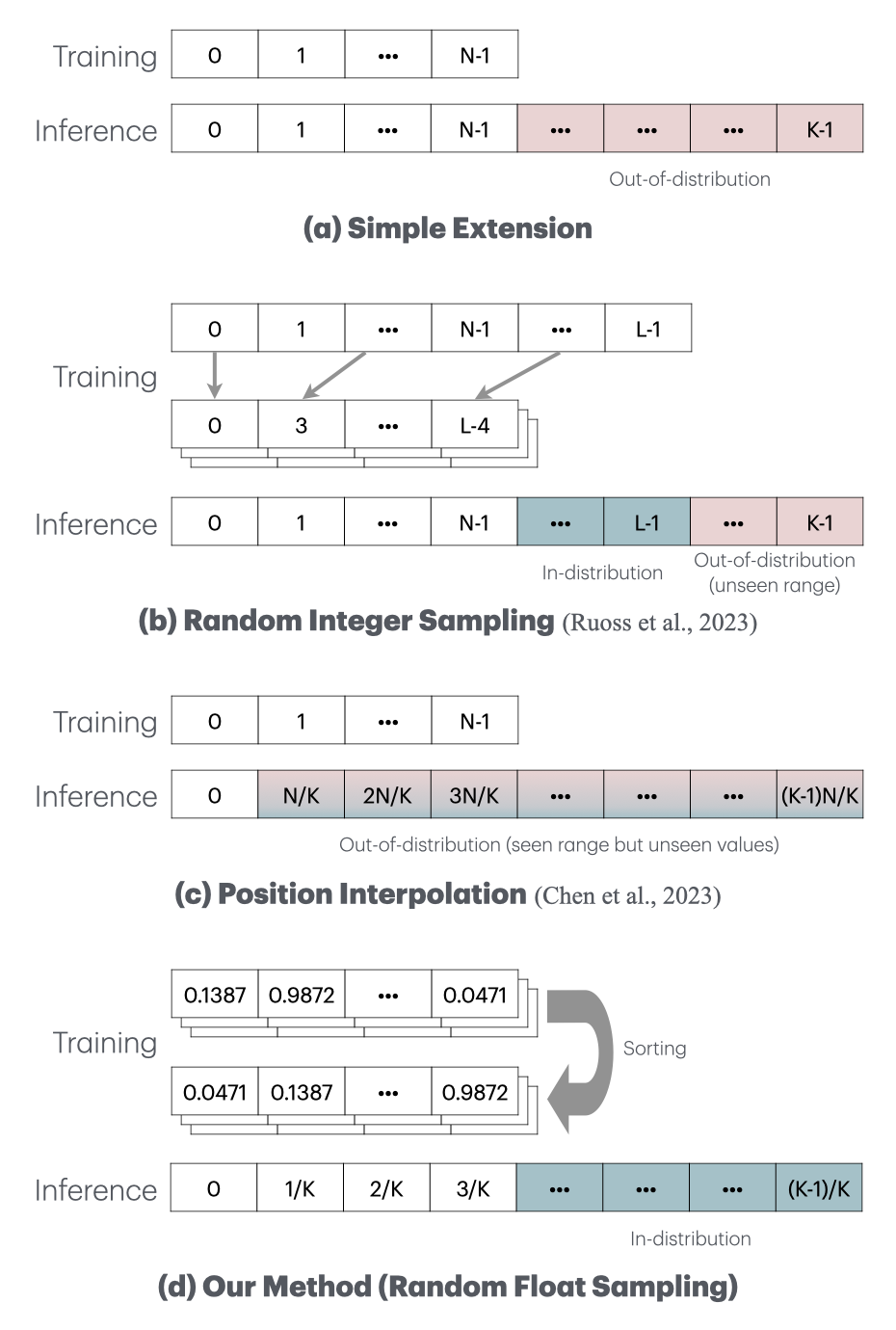}
    \caption{An illustration of the existing and proposed position indexing strategies. RFS avoids the OOD issue for any context length $K$, which is unknown during training.}
    \vspace{-\baselineskip}
    \label{fig:4methods}
\end{figure}

Length generalization is the problem of training language models that generalize to larger context sizes than those employed for pretraining. It has been a critical challenge because simply feeding longer contexts as in Figure \ref{fig:4methods} (a) causes an out-of-distribution (OOD) issue and thus fails, as shown empirically by \citet{press2021train, kazemnejad2023impact, zhou2024transformers}, while training on stretched sequences is costly, especially for Transformer-based models \cite{vaswani2017attention} due to their quadratic complexity. The need to handle longer contexts, however, has grown with advances in inference-time compute, such as prompt engineering \cite{kojima2022large}, Chain-of-Thought \cite{wei2022chain}, test-time scaling \cite{muennighoff2025s1simpletesttimescaling}, and the AI agents \cite{yao2023react, NEURIPS2023_1b44b878_reflexion}. The rise of small language models for real-world applications \cite{belcak2025small, liquid-ai} also necessitate the length generalization strategy as their context sizes are often smaller and prone to overflowing it, e.g., 2,048 for Phi-1.5 \cite{li2023textbooksneediiphi15}, TinyLlama \cite{zhang2024tinyllamaopensourcesmalllanguage}, and SmolLM2 \cite{allal2025smollm2smolgoesbig}.

Extensive approaches have been proposed to address this issue, including two notable works closely related to ours. \citet{ruoss-etal-2023-randomized} propose to sample $N$ integers from $\{1, \cdots, L\}$ uniformly at random for the position indices, given the maximum length $L$ and the training window size $N$ with $N < L$, as illustrated in Figure \ref{fig:4methods} (b). 
Position interpolation \cite{chen2023extendingcontextwindowlarge} down-scales the position index to match the original index range, and performs minimal fine-tuning with longer inputs, as shown in Figure \ref{fig:4methods} (c). Although both mitigate the issue by overlapping the position index range during training and inference, they still suffer from the OOD issue when the input length exceeds what was assumed during training or fine-tuning.

In this paper, we introduce a novel position indexing approach, Random Float Sampling (RFS), that technically allows the model to work on any input length by strategically selecting position indices from a shared continuous range during training and inference. Specifically, the indices are randomly sampled during training as depicted in Figure \ref{fig:4methods} (d), allowing the model to learn the attention mechanism based on relative token distances. Consequently, unlike existing PEs relying on absolute indices, the model trained under RFS is better at capturing the sequential structure even when an unexpected number of tokens are squeezed into the index range. 

\begin{figure}[t]
  \includegraphics[width=\columnwidth]{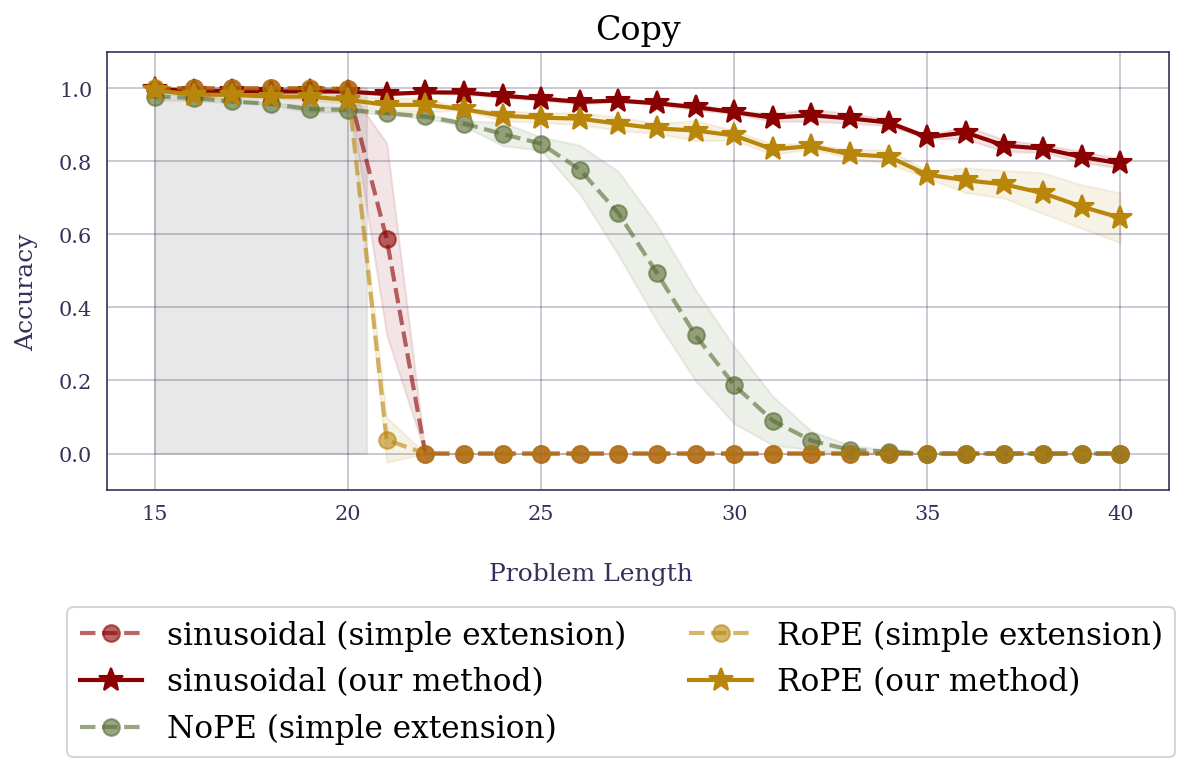}
  \caption{Results of different PEs on the copy task. The shaded area indicates that those lengths are seen during training. RFS boosts the performance significantly and leads to better results than NoPE. The experiment setup is stated in Section \ref{subsec:lg}.}
  \label{fig:copy3}
\end{figure}

Experiments on length generalization tasks demonstrate its effectiveness. As in Figure \ref{fig:copy3}, for example, with the absolute sinusoidal PE on the copy task, RFS maintains roughly $80\%$ accuracy on twice longer inputs, while the model without PE (NoPE), the best one in \citet{kazemnejad2023impact}, struggles on even 1.5 times longer inputs with accuracy around $20\%$. In another experiment on commonsense reasoning ability of language models, RFS shows a strong performance on unseen length test cases, which is comparable to that of a model without our method tested on seen length.

\section{Related Work}

\subsection{Position Encoding}

Transformers are permutation-equivariant; therefore, how to explicitly encode position information has been studied extensively. One common way is to form a matrix purely containing position information and add it to the token embedding matrix before they are fed into the Transformer blocks. \citet{vaswani2017attention} propose the use of sinusoids so that the dot product of two position vectors is roughly proportional to the token index distance. Another approach introduced in \citet{su2024roformer}, named RoPE, also utilizes sinusoidal functions and rotates queries and keys in each attention module to encode the token index distances. Since the above two PEs rely on sequential integer position indices, RFS can immediately replace them. 

\subsection{Length Generalization}

\noindent\textbf{RoPE-based architectures:} To train a model with long context length, \citet{dai2019transformer} and \citet{yu2023megabyte} propose architectures without full attention. Train-short-finetune-long is also considered in \citet{press2020shortformer, pmlr-v235-ding24i}, position interpolation \cite{chen2023extendingcontextwindowlarge}, and YaRN \cite{peng2023yarn}. Ours belongs to a train-short family that incurs no additional cost for fine-tuning. It includes position indexing strategies such as imposition of randomness for robustness, such as random integer sampling \cite{ruoss-etal-2023-randomized}, and others \cite{kiyono2021shape, likhomanenko2021cape}.

\noindent\textbf{Attention mechanism modification:} The T5 model \cite{raffel2020exploring} directly manipulates the attention scores with learnable parameters. This line of work includes ALiBi \cite{press2021train}, which is a lightweight version without learnable parameters. \citet{chi2022kerple, zheng2024dape, oka2025wavelet, nakanishi2025scalable} also propose improved attention mechanisms.

\noindent\textbf{No position encoding (NoPE):} Several studies cast doubt on the necessity of explicit positional information in causal Transformers \cite{haviv2022transformer, kazemnejad2023impact, irie2024positional}. They argue that position information can be learned and show that NoPE models perform as well as models with PE. Note that this only applies to causal models, not to bidirectional ones for translation \cite{vaswani2017attention} or transcription \cite{radford2023robust}.

\begin{figure*}[t]
  \includegraphics[width=\textwidth]{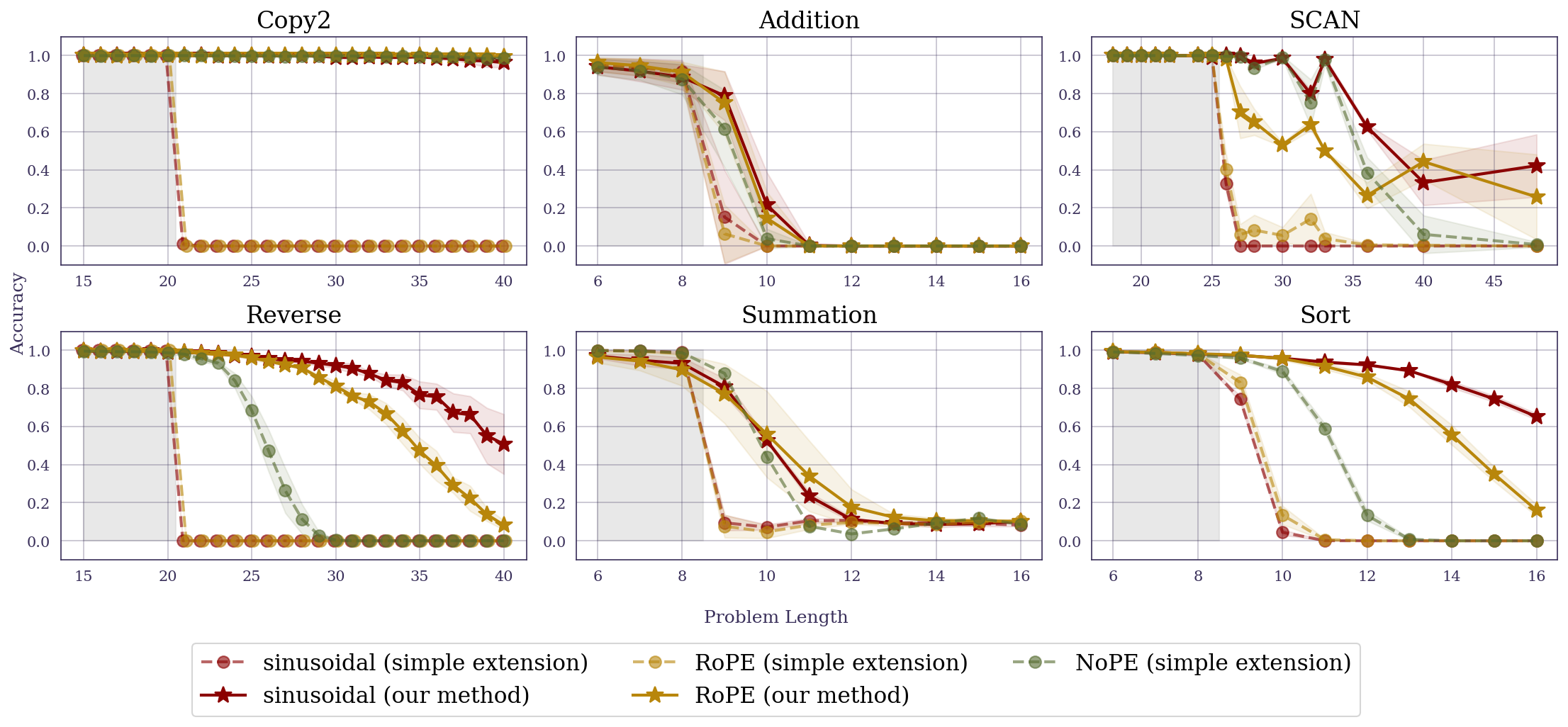}
  \caption{Results of length generalization tasks. The shaded area indicates the input lengths seen during training.}
  \label{fig:6tasks}
\end{figure*}

\section{Method: Random Float Sampling}\label{sec:method}

In this section, we formally define Random Float Sampling (RFS). Let $p_i$ be the position index of the $i$-th token and $[0, 1)$ be the continuous range where indices are chosen. 
During training, given a sequence of length $n_{\text{tr}}$, we sample $n_{\text{tr}}$ values from the range independently and uniformly at random, sort them in ascending order, and use them as the position indices. Namely, we have
\begin{align*}
    p_i = \tilde{p}_{\pi(i)}, \quad \tilde{p}_1, \tilde{p}_2, \cdots, \tilde{p}_{n_{\text{tr}}} \mathop\sim^{\text{i.i.d.}} [0, 1) ,
\end{align*}
where $\tilde{p}_{\pi(i)}$ is the $\pi(i)$-th smallest in $\tilde{p}_{1}, \cdots, \tilde{p}_{n_{\text{tr}}}$. 

In inference with a context of length $n_{\text{in}}$, the position indices are an evenly spaced partition of the shared range as 
\begin{align*}
    p_i =  \frac{\left( 2 i - 1 \right) }{2 \max(n_{\text{tr}}, n_{\text{in}})} .
\end{align*}
The indices are selected deterministically to make the model's output independent of the position index randomness. Note that when $n_{\text{in}} \ll n_{\text{tr}}$, the distances between the indices increase to a level nearly unseen during training. Using the max in the denominator prevents this degeneration.

Finally, the selected indices are scaled by $L$, a hyperparameter, to match the magnitude of indices commonly used in Transformers. We set it to $1,000$ in our experiments.
We again emphasize that once the position indices are obtained, they can immediately be plugged into the various PEs like absolute sinusoidal encoding, RoPE, and ALiBi.

\noindent\textbf{Design Rationale.} As position interpolation and random integer sampling imply, and as discussed in Appendix \ref{sec:whyfails}, we use the shared index range for training and inference. It immediately follows that, in inference, the token index distances vary depending on $n_{\text{in}}$. This is where randomness in training comes in, forcing the model to experience various position index distributions.

\section{Experiments}\label{experiments}

\subsection{Length Generalization Tasks}\label{subsec:lg}

\textbf{Setup.} To test our method's length generalization ability, we follow \citet{kazemnejad2023impact}, train a model from scratch on sequences up to a predefined length, and test on both seen and unseen lengths. The performance is evaluated by the exact-match accuracy on 6 tasks: copy and reverse \cite{ontanon-etal-2022-making}, addition \cite{nye2021workscratchpadsintermediatecomputation}, sort and summation \cite{saxton2019analysing}, and SCAN \cite{pmlr-v80-lake18a}. We refer readers to \citet{kazemnejad2023impact} and Appendix \ref{subsec:datasets}, \ref{subsec:lgt} for more details about the dataset.

\noindent\textbf{Architecture.} We use a decoder-only Transformer model with ``base'' configuration and the T5-Base tokenizer in the HuggingFace library \cite{wolf-etal-2020-transformers} and employ greedy decoding for generation. See Appendix \ref{subsec:lgt} for the full description. 
For position encoding, we consider the absolute sinusoidal PE and RoPE.
We compare our RFS with other position indexing, such as simple extension, position interpolation \cite{chen2023extendingcontextwindowlarge}, and random integer sampling \cite{ruoss-etal-2023-randomized}.
We do not fine-tune any model for fairness, even though \citet{chen2023extendingcontextwindowlarge} performs minimal fine-tuning in their experiments.

\begin{figure}[t]
  \includegraphics[width=\columnwidth]{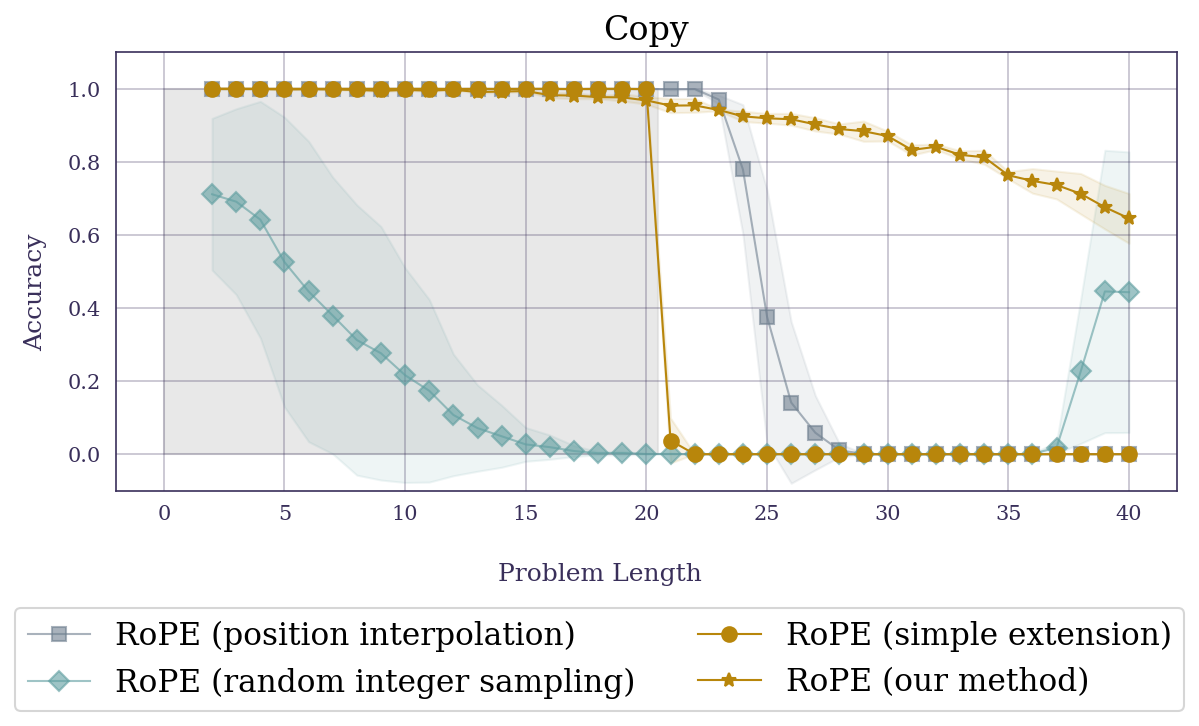}
  \caption{Results of RFS, position interpolation, and random integer sampling on the copy task. The shaded area indicates the input lengths seen during training.}
  \label{fig:rope_abcd}
\end{figure}

\noindent\textbf{Result.} We run the experiment with three different random seeds and report the performance in Figure \ref{fig:copy3} and \ref{fig:6tasks}. The most notable case is copy2; followed by the copy, reverse, and sort, where simple extension exhibits nearly no length generalization capabilities, but RFS still generates correct answers on unseen lengths. Second, RFS performs better than or at least on par with the NoPE, demonstrating the effectiveness of explicit position bias. Figure \ref{fig:rope_abcd} compares the RoPE-based RFS models with those of position interpolation and random integer sampling, highlighting the superiority of our method. It demonstrates that the advantage of position interpolation cannot be fully brought out without finetuning, while random integer sampling requires more training steps to converge as it underfits. Compared to random integer sampling, which also selects position indices from the predefined range, the use of floats in RFS not only relaxes the context size limitation but also accelerates the learning. Further discussion is given in Appendix \ref{subsec:pi_ris}. The result of ALiBi, which shows that RFS enhances the length generalization ability beyond absolute sinusoidal encoding and RoPE, is also relegated to Appendix \ref{subsec:alibi}.

\noindent\textbf{Ablation.} In Section \ref{sec:method}, RFS is defined as being trained with indices taken from the uniform distribution, and $L=1,000$ is employed. To test the alternative design choices, we run the experiment on the copy task using RoPE and report the average OOD accuracy. Firstly, the uniform distribution is replaced with the normal distribution with $0.5$ mean and $0.2$ standard deviation, or the beta distribution with $\alpha=0.5$ and $\beta=0.5$ for training, while inference-time indices follow those in Section \ref{sec:method}. Other variants with inference indices taken from the cumulative density functions of $[0, 1)$ even interpolation are also considered. Secondly, $L$ is modified to $10$ and $100,000$. The result, summarized in Table \ref{tab:ablation}, indicates that while simply supplanting the uniform distribution with the normal or beta distribution degrades the system, matching the inference indices distribution using the cumulative density functions performs well. Nevertheless, their OOD accuracy, $81.9\%$ and $81.2\%$ for the normal and beta distribution, respectively, falls short of the baseline $83.7\%$. The ablation on $L$ results in a slight degradation for both $L=10$ and $L=100,000$, showing that $L$ has a minimal impact on the OOD accuracy. Overall, these ablation studies justify the architectural choices of our RFS.

\begin{table}[t]
  \centering
  \begin{tabular}{cc}
    \toprule
    \textbf{Model} & \textbf{OOD accuracy} \\
    \midrule
    \textbf{Baseline} & \textbf{83.7} \\
    $\mathcal{N}(0.5, 0.2)$ & 0.0 \\
    $\mathcal{B}(0.5, 0.5)$ & 15.5 \\
    $\mathcal{N}(0.5, 0.2)$-cdf & 81.9 \\
    $\mathcal{B}(0.5, 0.5)$-cdf & 81.2 \\
    $L=10$ & 83.4 \\
    $L=100,000$ & 83.6 \\
    \bottomrule
  \end{tabular}
  \caption{Copy task performance over different design choices.}
  \label{tab:ablation}
\end{table}

\subsection{Language Modeling}

\begin{table*}
  \centering
  \resizebox{\textwidth}{!}{%
  \begin{tabular}{cccccccccc}
    \toprule
     &Task& \textbf{HellaSwag} & \textbf{RACE} & \textbf{ARC-e} & \textbf{ARC-c} & \textbf{OBQA} & \textbf{WinoGrande} & \textbf{BoolQ} & \textbf{Avg} \\
    & Context size & 87 & 439 & 28 & 33 & 13 & 21 & 130 \\
    \midrule
    \multirow{2}{*}{ID} & Baseline & \textbf{35.82} & 26.59 & 35.47 & \textbf{23.39} & \textbf{21.40} & 52.50 & \textbf{55.74} & 35.84 \\
     & Ours & 34.95 & \textbf{28.52} & \textbf{36.48} & 22.51 & 20.96 & \textbf{52.80} & 55.37 & \textbf{35.94} \\
    \midrule
    \multirow{2}{*}{OOD} & Baseline & \textbf{25.24} & \textbf{27.38} & \textbf{36.75} & 20.70 & 28.78 & 46.20 & 56.25 & 34.47 \\
     & Ours & 24.56 & 27.19 & 36.59 & \textbf{22.37} & \textbf{29.15} & \textbf{50.17} & \textbf{60.42} & \textbf{35.78} \\
    \bottomrule
  \end{tabular}
    }
  \caption{Zero-shot performance on commonsense reasoning tasks. ARC-e, ARC-c, and OBQA are short for ARC-Easy, ARC-Challenge, and OpenbookQA, respectively.}
  \label{tab:lm_result}
\end{table*}

\noindent\textbf{Setup.} To evaluate the effectiveness of our method in language modeling, we train language models from scratch on OpenWebText \cite{Gokaslan2019OpenWeb} and test their zero-shot commonsense reasoning ability on the following 7 tasks: HellaSwag \cite{zellers2019hellaswag}, RACE \cite{lai-etal-2017-race}, ARC-Easy and ARC-Challenge \cite{Clark2018ThinkYH_arc}, OpenBookQA \cite{OpenBookQA2018}, WinoGrande \cite{sakaguchi2019winogrande}, and BoolQ \cite{boolq}. Additional details about the dataset are provided in Appendix \ref{subsec:datasets}. 
To examine the length generalization ability, we set the model’s context window to the median number of tokens in each task’s test set, splitting the set into in-distribution (ID) and out-of-distribution (OOD) subsets.

\noindent\textbf{Architecture.} The models are based on GPT-2 \cite{radford2019language} with 12 layers, each with 12 heads, and an embedding dimension of 768. 
We use the same tokenizer with GPT-2.
Details are summarized in Appendix \ref{sec:lm_detail}.

\noindent\textbf{Result.} Table \ref{tab:lm_result} reports the result. Our RFS showcases the better average accuracy in the OOD test cases, while performing on par with the baseline in the ID cases. Notably, the average accuracy of our method in OOD $35.78\%$ is comparable to that of the baseline in ID tests $35.84\%$, highlighting the length generalization ability of our PE strategy. Namely, the degraded performance of the baseline on OOD samples, which stems from the lack of length generalization ability, is almost recovered by RFS.
Although RFS does not improve OOD accuracy on HellaSwag, RACE, and ARC-Easy, this may come from training data or architectural factors rather than indexing itself; we leave a deeper analysis to future work.

\section{Conclusion}

In this paper, we introduce Random Float Sampling, a random position indexing strategy in Transformers that places no intrinsic upper bound on context length at the indexing level. RFS is simple enough to be integrated into many commonly used PEs, and its advantage is evaluated from various angles. RFS demonstrates the strongest performance against existing methods on length generalization tasks with a significant margin. It also improves the accuracy on OOD samples in commonsense reasoning tasks to nearly the same level as on their ID counterparts, showcasing the practicality of RFS in language modeling.

\section*{Limitations}

Firstly, we conducted the experiments on language models with around 100 million parameters, but not on the commercial-size LLMs, due to computational resource constraints. To fully discover its potential in real-world applications, testing on larger LLMs is necessary. Secondly, even though our method technically accepts contexts of any length, its performance deteriorates as the input becomes longer. To compensate for degradation, combining random position indexing with other orthogonal works, such as Scalable-Softmax \cite{nakanishi2025scalable} and interleaving deployment of RoPE and NoPE \cite{yang2025ropenopeagainnew}, would be an interesting direction for future research.


\section*{Acknowledgments}

This research was conducted using the FUJITSU Supercomputer PRIMEHPC FX1000 and FUJITSU Server PRIMERGY GX2570 (Wisteria/BDEC-01) at the Information Technology Center, the University of Tokyo.

\bibliography{bibliography}

\clearpage

\appendix

\section{Why Does Simple Extension Fail?}\label{sec:whyfails}

Let $\mathbf{x} \in \mathbb{R}^d$ be the vector representation of a token, which is then fed into the attention block and projected to a query and a key by computing $\mathbf{q} = f_q(\mathbf{x})$ and $\mathbf{k} = f_k(\mathbf{x})$ where $f_q$ and $f_k$ represent the linear transformation and rotary positional encoding (RoPE). \citet{chen2023extendingcontextwindowlarge} shows that, given position differences, there always exist $f_q$ and $f_k$ that produce well-behaved attention scores $\mathbf{q}^T \mathbf{k}$, but the attention scores can explode for unseen position differences, which ruin the self-attention mechanism. They also proved that attention scores behave nicely under position interpolation, where position differences can be unseen but within the range seen during training. Their proof can be applied to our RFS as well, guaranteeing the attention scores won't spike.

\begin{figure}[t]
    \centering
    \begin{subfigure}[b]{0.49\textwidth}
        \centering
        \includegraphics[width=\textwidth]{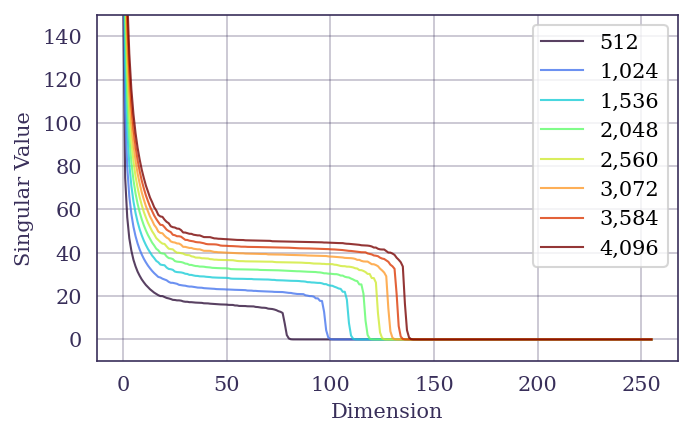}
        \caption{Simple Extension}
    \end{subfigure}
    \hfill
    \begin{subfigure}[b]{0.49\textwidth}
        \centering
        \includegraphics[width=\textwidth]{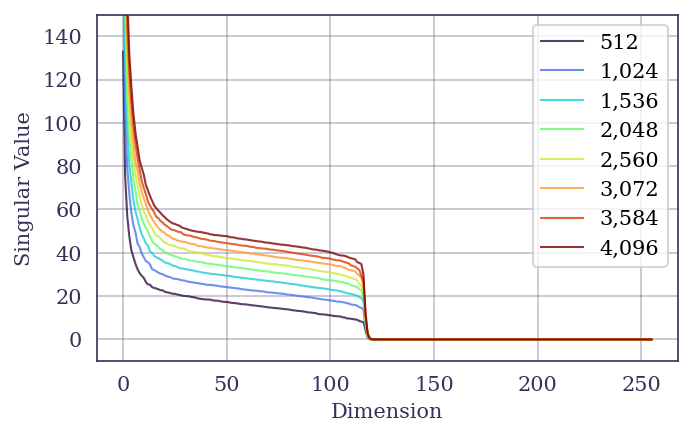}
        \caption{Our Method (RFS)}
    \end{subfigure}
    \caption{The singular values of the additive sinusoidal position matrix. The dimensionality is 256. The position index range is set to $[0, 2048]$ in our method (bottom). With the simple extension, the longer the input gets, the higher the rank of the position matrix becomes. RFS effectively controls the rank no matter the input length.}
    \label{fig:rank}
\end{figure}

For the absolute sinusoidal encoding, let $\mathbf{P} \in \mathbb{R}^{n \times d}$ be the position matrix where $i$-th row $\mathbf{p}_i$ is the position vector defined as  
\begin{align}\label{eq:sinusoidal}
    \begin{split}
        \mathbf{p}_{i, 2k} &= \sin(p_i / 10000^{2k / d}) \\
        \mathbf{p}_{i, 2k+1} &= \cos(p_i / 10000^{2k / d}).
    \end{split}
\end{align}
We argue that the simple extension fails because the rank of the position matrix increases as the input sequence gets longer, interfering with the subspace purely for the semantic information during training. Figure \ref{fig:rank} clearly illustrates this, where we perform SVD on the position matrix and plot the singular values in descending order. We set $d=256$ and choose 8 values for $n$. Note that the rank of a matrix is the same as the number of its non-zero singular values, and no position matrix in Figure \ref{fig:rank} is a full rank. Considering that the position matrix is added to the token embedding $\mathbf{X} \in \mathbb{R}^{n \times d}$, the input to the self-attention block $\mathbf{X} + \mathbf{P}$ stores the position information only in the smaller subspace, and the rest only holds the semantic information. We observe that the simple extension increases the rank of the position matrix as the sequence length grows, which is problematic as it undermines the subspace in which the model assumes no position information is contained. This is not the case for our RFS.

\noindent\textbf{Takeaway.} In both RoPE and the absolute sinusoidal encoding, utilizing unseen position indices during training ruins the model because they are out-of-distribution, and random integer sampling and position interpolation overcome this issue by matching the position index range in training and inference. Thus, squeezing the indices into the range $[0, L]$ is the natural starting point for us. Further, inspired by the observation that position indices used in the inference must be seen during training, we design our position indexing strategy so that the model is forced to experience not only various position indices within the range but also diversified position distances where the extra randomness comes in. This differentiates our method from position interpolation and leads to the distinctive performance as demonstrated in Section \ref{experiments}.

\section{Experiment Details}\label{sec:lg_detail}

\subsection{Code}
The code for this research is available at \href{https://anonymous.4open.science/r/RFS-arr-submission/}{https://anonymous.4open.science/r/RFS-arr-submission/}.

\subsection{Datasets}\label{subsec:datasets}
Table \ref{tab:datasets} provides a summary of datasets used in the experiments.

\begin{table*}
  \centering
  \begin{tabular}{llrrl}
    \toprule
    \textbf{Name} & \textbf{Split} &  \textbf{\# training samples} & \textbf{\# test samples} & \textbf{License} \\
    \midrule
    Copy & cmc\_tr20\_ts40 & 100,000 & 10,000 & MIT License \\
    Copy2 & rsc\_tr20\_ts40 & 100,000 & 10,000 & MIT License \\
    Reverse & mc\_tr20\_ts40 & 100,000 & 10,000 & MIT License \\
    Addition & len\_tr8\_ts16 & 100,000 & 10,000 & MIT License \\
    Sort & len\_mltd\_tr8\_ts16 & 100,000 & 10,000 & MIT License \\
    Summation & len\_tr8\_ts16 & 100,000 & 10,000 & MIT License \\
    SCAN & len\_tr25\_ts48 & 15,997 & 3,136 & BSD License \\
    OpenWebText & - & - & - & Creative Commons \\
    HellaSwag & - & - & 10,042 & MIT License \\
    RACE & - & - & 1,045 & - \\
    ARC-e & - & - & 2,376 & Creative Commons \\
    ARC-c & - & - & 1,172 & Creative Commons \\
    OpenbookQA & - & - & 500 & - \\
    WinoGrande & - & - & 1,267 & - \\
    BoolQ & - & - & 3,270 & Creative Commons \\
    \bottomrule
  \end{tabular}
  \caption{Dataset description.}
  \label{tab:datasets}
\end{table*}

\begin{table*}
  \centering
  \begin{tabular}{lll}
    \toprule
    \textbf{Task Name} & \textbf{Input Example} & \textbf{Output Example} \\
    \midrule
    Copy & Copy: Code Bank southern 8 & Code Bank southern 8 \\
    Copy2 & Copy: Angle Angle Angle & Angle Angle Angle \\
    Reverse & situated Con branches 8 & 8 branches Con situated \\
    Addition & Compute 3 5 3 2 + 6 2 7 6 5 4 ? & The answer is 6 3 1 1 8 6. \\
    Sort & Sort the following list [ 9 8 3 0, 5 1 5 ].  & The sorted list is [ 5 1 5, 9 8 3 0 ]. \\
    Summation & Compute: 3 + 9 + 2 + 1 + 7 . & The answer is 2 . \\
    SCAN & jump twice after run right & Right Run Jump Jump \\
    \bottomrule
  \end{tabular}
  \caption{Examples of the input and output of the length generalization tasks.}
  \label{tab:lg_tasks}
\end{table*}

\subsection{Computational Resources}
All model training is performed on eight Nvidia A100 (40GB) instances. For the length generalization tasks, one training run is completed in 4 GPU hours. For the language modeling, one training run is completed in 160 GPU hours.

\subsection{Length Generalization Tasks}\label{subsec:lgt}

In Section \ref{subsec:lg}, we evaluated our method on 7 tasks: copy, copy2, reverse, addition, sort, summation, and SCAN. The examples of them are provided in Table \ref{tab:lg_tasks}. Note that we ask to answer only one's place of the result in the summation task following \citet{kazemnejad2023impact}. Table \ref{tab:lg_params} presents the model architecture and the training parameters. A hyperparameter $L$ in our method, the range from which position indices are selected, is set to $1,000$ for absolute sinusoidal encoding and RoPE, while for ALiBi, we select the maximum input length of the entire dataset.

\begin{table}[h]
  \centering
  \resizebox{\columnwidth}{!}{%
  \begin{tabular}{lr}
    \toprule
    \textbf{Parameter} & \textbf{Value} \\
    \midrule\midrule
    \multicolumn{2}{c}{Architecture} \\
    \midrule
    \# Layers & 12 \\
    \# Attention Heads & 12 \\
    Model Dimension & 768 \\
    Rotary Dimension & 16 (25\% of the head dim) \\
    Normalization & LayerNorm \\
    Dropout & 0.1 \\
    Decoding Strategy & Greedy \\
    \midrule
    \multicolumn{2}{c}{Hyperparameters} \\
    \midrule
    Optimizer & AdamW \\
    Learning Rate & 3e-5 \\
    Weight Decay & 0.05 \\
    $\beta_1$ & 0.9 \\
    $\beta_2$ & 0.999 \\
    Gradient Clipping & 1.0 \\
    Batch Size & 64 \\
    \# Train Steps & 40,000 \\
    \# Warm Ups & 2,400 \\
    Index Range $L$ (ALiBi) & max input length \\
    Index Range $L$ (others) & 1,000 \\
    \bottomrule
  \end{tabular}
  }
  \caption{Summary of model architecture and the training parameters for the length generalization tasks.}
  \label{tab:lg_params}
\end{table}

\subsection{Language Modeling Experiment Details}\label{sec:lm_detail}

The architecture and the hyperparameters are detailed in Table \ref{tab:lm_params}. After being pretrained on OpenWebText \cite{Gokaslan2019OpenWeb}, the models are evaluated using the Language Model Evaluation Harness \cite{eval-harness}.

\begin{table}[h]
  \centering
  \begin{tabular}{lr}
    \toprule
    \textbf{Parameter} & \textbf{Value} \\
    \midrule\midrule
    \multicolumn{2}{c}{Architecture} \\
    \midrule
    \# Layers & 12 \\
    \# Attention Heads & 12 \\
    Model Dimension & 768 \\
    Position encoding & RoPE \\
    Normalization & LayerNorm \\
    Dropout & 0.0 \\
    \midrule
    \multicolumn{2}{c}{Hyperparameters} \\
    \midrule
    Optimizer & AdamW \\
    Learning Rate & 6e-4 \\
    Weight Decay & 0.05 \\
    $\beta_1$ & 0.9 \\
    $\beta_2$ & 0.999 \\
    Gradient Clipping & 1.0 \\
    Batch Size & 480 \\
    \# Train Steps & 100,000 \\
    \# Warm Ups & 2,000 \\
    Index Range $L$ & 1,000 \\
    \bottomrule
  \end{tabular}
  \caption{Summary of model architecture and the training parameters for the language modeling.}
  \label{tab:lm_params}
\end{table}

\section{Length Generalization Tasks Additional Results}\label{sec:lg_add_result}

This section provides a qualitative error analysis and two additional experiments to further clarify the advantages of the proposed method. The first experiment compares RFS to the performance of position interpolation and random integer sampling on the length generalization tasks. Then, an experiment of testing the compatibility of ALiBi and our method follows. The experimental settings are identical to those in Section \ref{subsec:lg}. 

\subsection{Error Analysis}\label{subsec:error}
Table \ref{tab:error} reports validation set samples that both the baseline and RFS with RoPE predict incorrectly. A sample of the copy2 task is not included, as RFS made no errors. Though incorrect, these are not completely nonsense, and RFS answers tend to be closer to the target. For instance, in the copy task, RFS prediction has only one word error, while the baseline additionally skips two words. Another example in the sort task, RFS answers only one number incorrectly, but the baseline additionally forgets to generate a comma. It suggests that when other metrics other than the exact match are employed, the performance gap between the baseline and RFS may get larger. This motivates practitioners to equip RFS in a situation where minor errors are acceptable.

\begin{table*}
  \centering
  \resizebox{\textwidth}{!}{
  \begin{tabular}{ll}
    \toprule
    \multicolumn{2}{c}{Copy | Copy: 1 southern I M J \$ high Hersteller conform J B J F Positive heroic Men 3 A + 4 registered} \\
    \midrule
    Target & 1 southern I M J \$ high Hersteller conform J B J F Positive heroic Men 3 A + 4 registered \\
    Baseline & 1 southern I M J \$ high Hersteller conform J F J F Positive heroic Men 3 A registered \\
    RFS & 1 southern I M J \$ high Hersteller conform J F J F Positive heroic Men 3 A + 4 registered \\
    \midrule\midrule
    \multicolumn{2}{c}{Reverse | 2 tears tears S neighborhoods Cover = geschaffen C apa tears 8 5 exp R 2 despair E ( 5 A branches 1} \\
    \midrule
    Target & 1 branches A 5 ( E despair 2 R exp 5 8 tears apa C geschaffen = Cover neighborhoods S tears tears 2 \\
    Baseline & E 2 2 R 8 8 tears apa C geschaffen geschaffen = Cover neighborhoods neighborhoods tears tears 2 2 2 \\
    RFS & 1 branches A 5 ( E despair 2 R exp 5 8 tears apa C geschaffen = Cover neighborhoods S tears 2 2 \\
    \midrule\midrule
    \multicolumn{2}{c}{Addition | Compute 5 0 + 6 0 1 8 7 7 3 1 1 0 5 ?} \\
    \midrule
    Target & 60187731155 \\
    Baseline & 6 0 1 8 7 7 6 5. 5. 5. 5. \\
    RFS & 6 0 1 8 7 7 3 1 5 5. 5. \\
    \midrule\midrule
    \multicolumn{2}{c}{Sort | Sort the following list [ 9 7 6 6, 4 3 5 9, 4 2 2 9, 1 4 4 4, 6 4 6 3, 7 1 4 5, 4 7 5 4, 4 5 5, 4 1 3 9, 4 0 4 5 ].} \\
    \midrule
    Target & 455, 1444, 4045, 4139, 4229, 4359, 4754, 6463, 7145, 9766 \\
    Baseline & 4 5 5, 1 4 4 4, 4 0 4 5, 4 1 3 9, 4 2 2 9, 4 3 5 9, 6 7 5 4, 6 4 6 3, 7 1 4 5 9 7 6 6 \\
    RFS & 4 5 5, 1 4 4 4, 4 0 4 5, 4 1 3 9, 4 2 2 9, 4 7 5 9, 4 7 5 4, 6 4 6 3, 7 1 4 5, 9 7 6 6  \\
    \midrule\midrule
    \multicolumn{2}{c}{Summation | Compute: 2 + 4 + 7 + 9 + 9 + 4 + 2 + 9 + 2 .} \\
    \midrule
    Target & 8 \\
    Baseline & 6 \\
    RFS & 9  \\
    \midrule\midrule
    \multicolumn{2}{c}{SCAN | look around right thrice and turn opposite right} \\
    \midrule
    \multirow{2}{*}{Target} & Right Look Right Look Right Look Right Look Right Look Right Look Right Look Right Look \\ & Right Look Right Look Right Look Right Look Right Right \\
    \multirow{2}{*}{Baseline} & Right Look Right Look Right Look Right Look Right Look Right Look Right Look Right Look \\ & Right Look Right Look Right Look Right Right Right Right \\
    \multirow{2}{*}{RFS} & Right Look Right Look Right Look Right Look Right Look Right Look Right Look Right Look \\ & Right Look Right Look Right Look Right Right Right Right \\
    \bottomrule
  \end{tabular}
  }
  \caption{Typical errors on algorithmic tasks.}
  \label{tab:error}
\end{table*}

\subsection{Position Interpolation and Random Integer Sampling}\label{subsec:pi_ris}
For position interpolation, we evaluate the performance without fine-tuning for fair comparison, even though \citet{chen2023extendingcontextwindowlarge} states that minimal fine-tuning is necessary to fully leverage its architectural edge. In random integer sampling, the maximum length $L$ is $512$. We run the experiment 3 times for each setup and take the average accuracy for visualization.

\begin{figure*}[t]
    \centering
    \begin{subfigure}[b]{\textwidth}
        \centering
        \includegraphics[width=\textwidth]{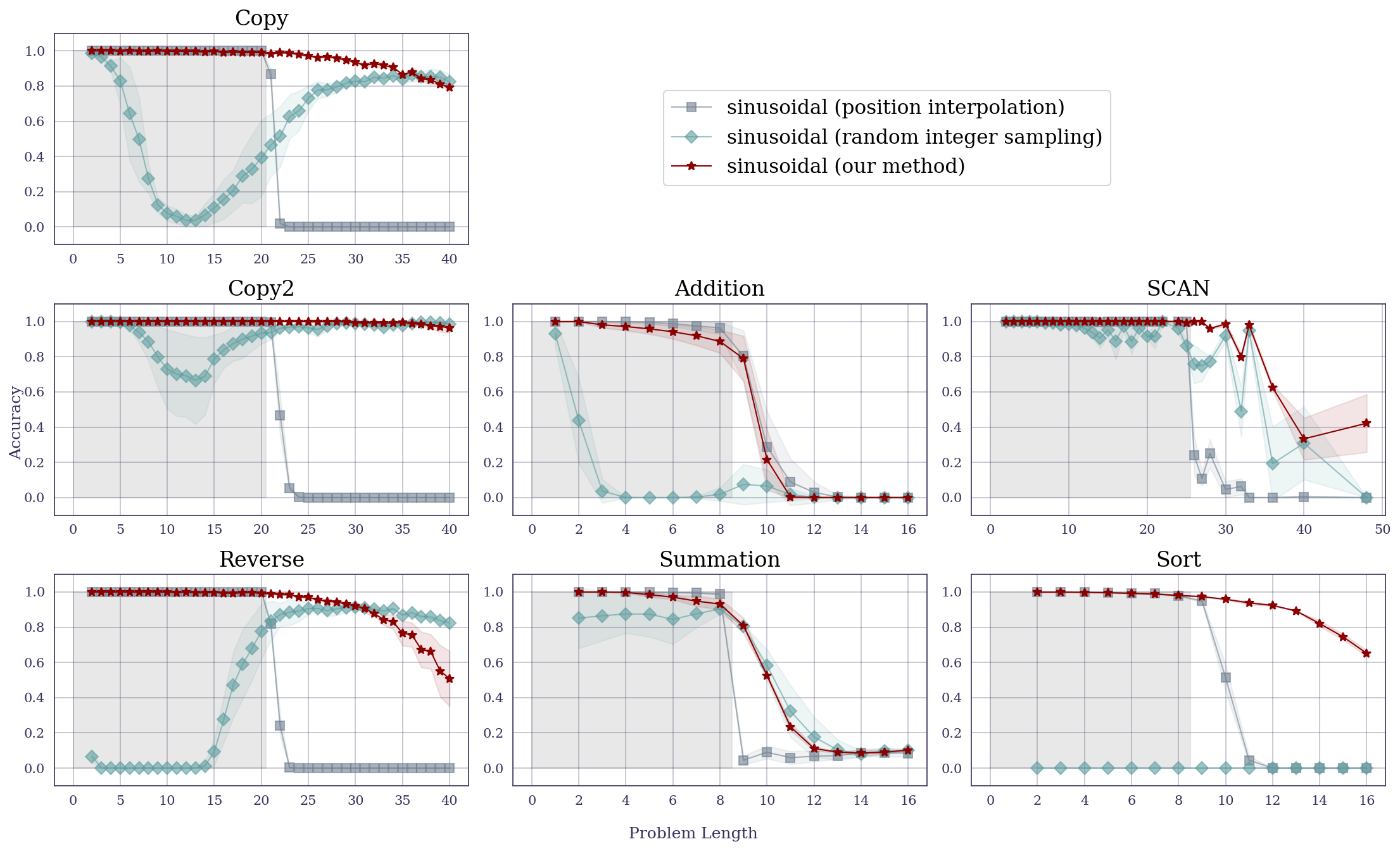}
        \caption{Sinusoidal}
    \end{subfigure}
    \\
    \begin{subfigure}[b]{\textwidth}
        \centering
        \includegraphics[width=\textwidth]{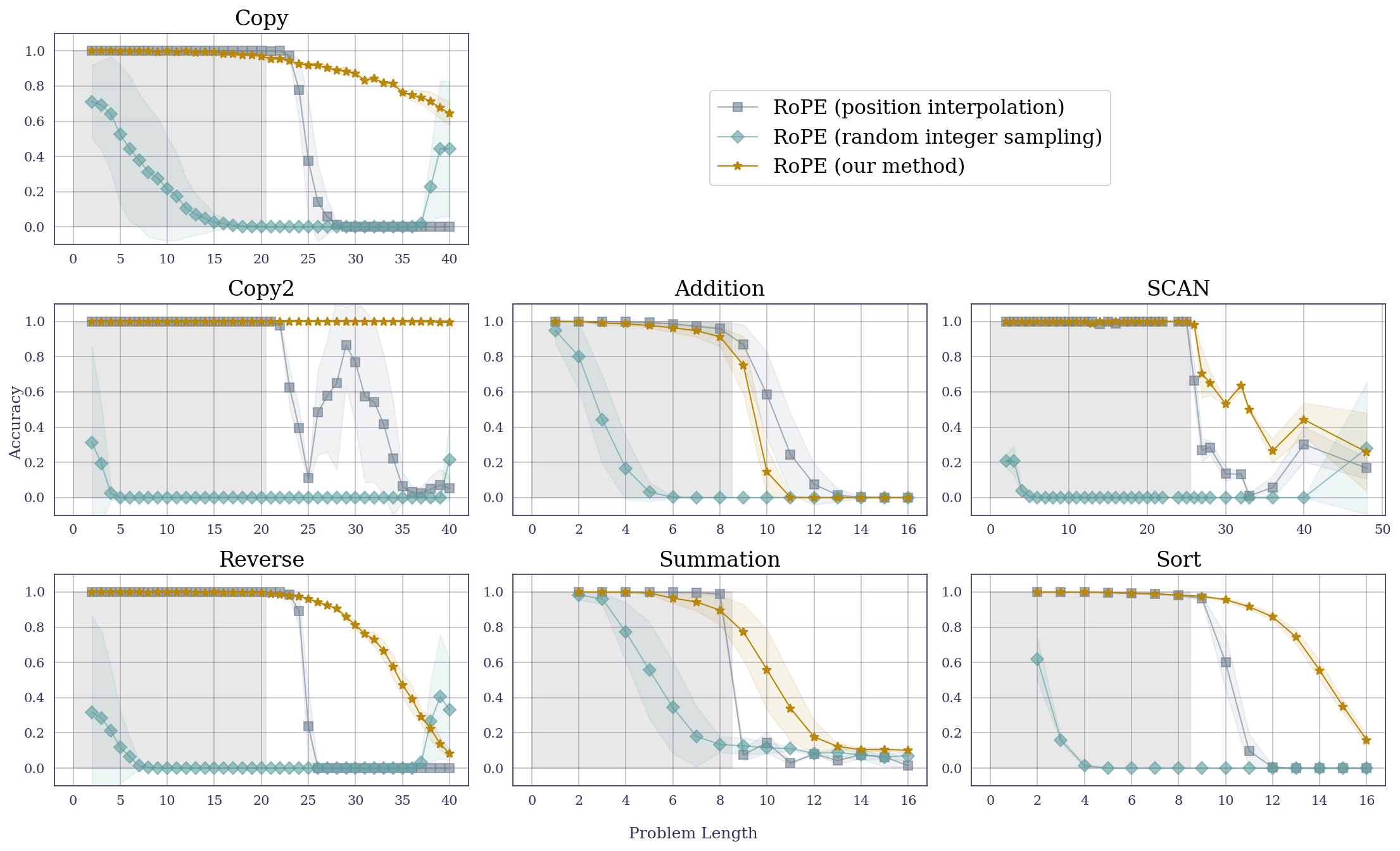}
        \caption{RoPE}
    \end{subfigure}
    \caption{The results of position interpolation without fine-tuning, random integer sampling, and our method on the length generalization tasks.}
    \label{fig:7tasks}
\end{figure*}

The results are given in Figure \ref{fig:7tasks}. Overall, our method presents better length generalization ability in most of the tasks, although for the addition task, position interpolation exhibits a slightly better performance than ours. Random integer sampling often struggles to achieve high accuracy for problem lengths seen during training, implying that this method underfits and requires more training steps.

\subsection{Random Float Sampling and ALiBi}\label{subsec:alibi}
ALiBi is designed to enhance the length generalization ability by not modifying the queries and keys in the attention mechanism but adjusting the attention scores according to the token distances before the softmax operator. Thus, it avoids the OOD issues discussed in \ref{sec:whyfails} that are inherent to the absolute sinusoidal encoding and RoPE. However, attention score adjustments can be unseen if the input length exceeds the training context size, and it may cause performance degradation, which could be mitigated by incorporating our method.

\begin{figure*}[t]
    \begin{subfigure}[b]{\textwidth}
        \centering
        \includegraphics[width=\textwidth]{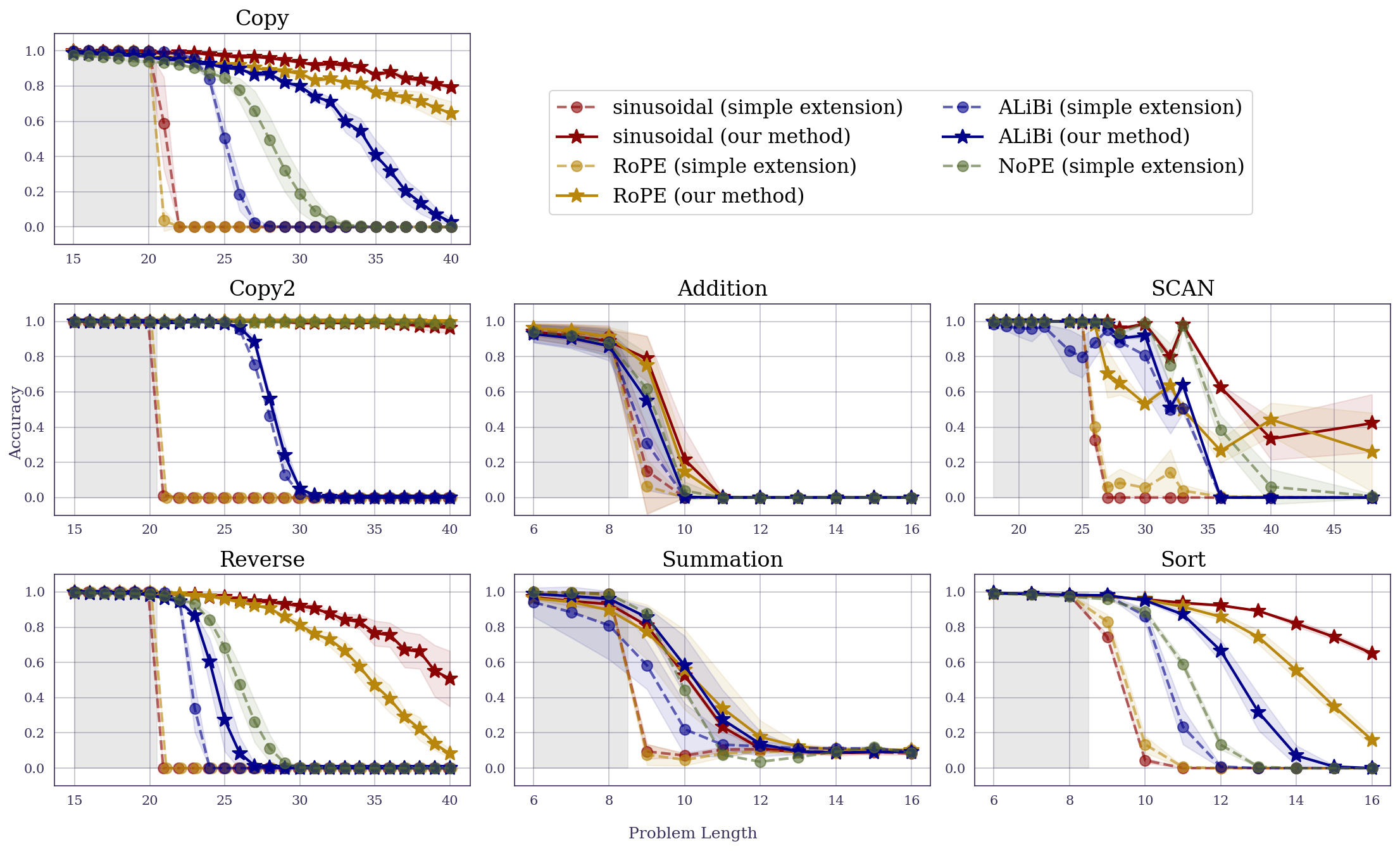}
    \end{subfigure}
    \caption{The impact of the proposed method on ALiBi (blue lines) on the length generalization tasks.}
    \label{fig:alibi}
\end{figure*}

This hypothesis is tested and reported in Figure \ref{fig:alibi} by adding two lines, ALiBi with our method and with the simple extension, to Figure \ref{fig:6tasks}. It clarifies that our method successfully adds performance gain to the simple extension in all the tasks, suggesting that experiencing possible attention score adjustments during training leads to better length generalization ability. That being said, the performance gain is smaller than that in the absolute sinusoidal encoding and RoPE, and it shows limited effectiveness than NoPE in copy2, reverse, addition, and SCAN.

\section{Use of LLM}
We employed a large language model (ChatGPT; “GPT-5 Thinking”) only for English‐language polishing and light copy-editing. 
The model was not used to generate ideas and experimental design. 
All technical content and claims were authored and verified by the human authors. 
No non-public data, confidential information, or personally identifiable information were provided to the model. 
\end{document}